\documentclass[letterpaper, 10 pt, conference]{ieeeconf}
\IEEEoverridecommandlockouts
\usepackage{graphicx} %
\usepackage[dvipsnames]{xcolor}
\usepackage{xspace}
\usepackage{amsmath}
\usepackage{amsfonts}
\usepackage{subcaption}
\usepackage{hyperref}
\usepackage{multirow}
\usepackage{hhline}

\usepackage[letterpaper, lmargin=54pt, tmargin=54pt, rmargin=54pt, bmargin=57pt]{geometry}

\makeatletter
\DeclareRobustCommand\onedot{\futurelet\@let@token\@onedot}
\def\@onedot{\ifx\@let@token.\else.\null\fi\xspace}

\def\etal{\emph{et al}\onedot}
\makeatother

\title{\LARGE \bf
Topologically-Informed Atlas Learning
}

\author{Thomas Cohn$^{1}$, Nikhil Devraj$^{1}$, Odest Chadwicke Jenkins$^{1}$%
\thanks{$^{1}$Thomas Cohn, Nikhil Devraj, and Odest Chadwicke Jenkins are with the Department of Electrical Engineering and Computer Science Engineering, Robotics Institute,  University of Michigan,
        Ann Arbor, MI 48109, USA
        {\tt\small [cohnt|devrajn|ocj]@umich.edu}}%
}

\begin{document}

\maketitle
\thispagestyle{empty}
\pagestyle{empty}

\begin{abstract}%
    We present a new technique that enables manifold learning to accurately embed data manifolds that contain holes, without discarding any topological information.
Manifold learning aims to embed high-dimensional data into a lower dimensional Euclidean space by learning a coordinate chart, but it requires that the entire manifold can be embedded in a single chart.
This is impossible for manifolds with holes.
In such cases, it is necessary to learn an atlas: a collection of charts that collectively cover the entire manifold.
We begin with many small charts, and combine them in a bottom-up approach, where charts are only combined if doing so will not introduce problematic topological features.
When it is no longer possible to combine any charts, each chart is individually embedded with standard manifold learning techniques, completing the construction of the atlas.
We show the efficacy of our method by constructing atlases for challenging synthetic manifolds; learning human motion embeddings from motion capture data; and learning kinematic models of articulated objects.
\end{abstract}

\section{Introduction}%
\label{sec:introduction}

The data available to roboticists are rapidly improving in quality and quantity.
We are seeing extensive improvements such as more detailed object models, higher fidelity sensor readings, and greater diversity in modality. %
Although such data offer incredible opportunities in the field of robotics, they also come with challenges inherent to their high dimensionality. %
For example, many machine learning algorithms quickly become ineffective or intractable in high-dimensional spaces~\cite{aggarwal2001surprising,snyder2008obstacles,marron2007distance}. %
The sparsity of a dataset grows exponentially with dimension, and it becomes more difficult to detect outliers and handle the bias-variance tradeoff~\cite{bias_variance_nn}.
Interfacing with humans for data visualization and robot teleoperation quickly becomes infeasible.
This ``curse of dimensionality'' is a prevalent challenge in robotics, so techniques designed to handle high-dimensional data are of great importance.

One appealing technique for handling high-dimensional data is \textit{dimensionality reduction}: constructing a new representation of the original data in a lower dimensional space, while still preserving the latent information.
Manifold learning is a class of dimensionality reduction algorithms which model the data as lying along some smooth manifold embedded in the high-dimensional space.
These algorithms construct a mapping from the manifold to a lower-dimensional Euclidean space, in effect, ``unwrapping'' the manifold.
Manifold learning has been applied to learn useful data representations in many areas of robotics~\cite{jenkins2002deriving,romero2013extracting,sinapov2008detecting}.

A drawback of manifold learning is the implicit assumption that the data can be unwrapped in one piece.
Manifolds need only be locally Euclidean: around any point, there is a small region which can be flattened.
But globally, manifolds may contain topological holes (such as the space enclosed inside a sphere and the inner loop of a torus), which make it impossible to unwrap with a single chart.

\begin{figure}[t]
    \centering
    \begin{subfigure}[t]{0.58\linewidth}
        \includegraphics[width=\linewidth]{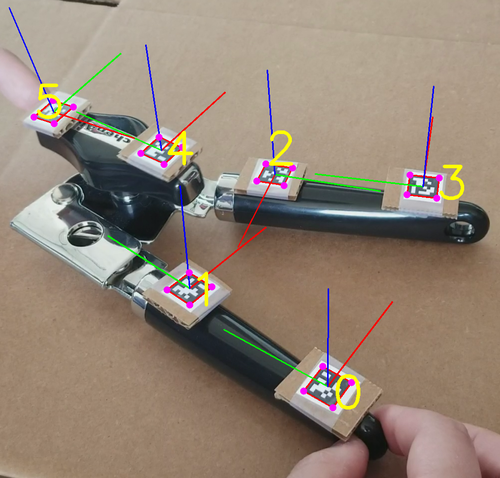}
        \caption{}
    \end{subfigure}
    \hfill
    \begin{subfigure}[t]{0.28\linewidth}
        \centering
        \includegraphics[width=\linewidth]{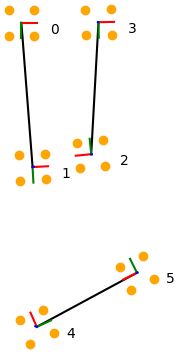}
        \caption{}
    \end{subfigure}
    \caption{This figure shows part of an experiment involving the construction of kinematic models for articulated objects. (a) is an input image of a can opener, with added AprilTag fiducial markers~\cite{apriltags} to obtain link poses. Each tag is annotated with its computed pose, as well as its numerical identifier. (b) is a new object state, obtained from the learned low-dimensional kinematic model (produced by our atlas learning technique), in the form of tag corner locations and poses.}
    \label{fig:teaser}
\end{figure}

It is possible to identify holes, and make cuts to the manifold so that it can be flattened out, as was done in CycleCut~\cite{cyclecut}.
But to handle these cases without losing connectivity information, an extension to manifold learning, called \textit{atlas learning}, is needed.
The mathematical definition of a manifold is a collection of overlapping coordinate charts, called an \textit{atlas}.
Atlas learning constructs such a family of charts, as opposed to just a single mapping.
This representation can preserve the global topology of any manifold, while still admitting low-dimensional representations.
Most previous atlas approaches have restricted their charts to be linear projections, to admit fast, simultaneous optimization of domain assignments and the learned
embedding~\cite{learning_a_manifold_as_an_atlas,unsupervised_atlas}, or to simplify construction of the atlas from an intermediate representation~\cite{manifold_tangent_classifier}.
But linear mappings must be restricted to a small domain to accurately represent nonlinear data, so these atlas approaches require a large number of charts.
An autoencoder approach~\cite{korman_atlas} can construct nonlinear charts, but the charts may not preserve their domain's topology.

In this paper, we present a new, topologically-informed approach to atlas learning.
The manifold is initially divided into many small patches, and then we combine patches in a bottom-up approach, provided that doing so will not lead to the patch containing a hole.
We present an efficient technique for checking when patches can be combined, informed by the theory of de Rham cohomology on smooth manifolds~\cite[Ch.~7]{manifolds_book} and leveraging the work of CycleCut to detect topological holes in graphs~\cite{cyclecut}.
When no more patches can be combined, they are individually embedded using ISOMAP~\cite{isomap}.
We demonstrate the efficacy of our approach by learning atlases of computer-generated synthetic manifolds, embedding human motion capture data, and learning kinematic models of articulated objects.

\section{Related Work}%
\label{sec:related_work}

Most manifold learning algorithms~\cite{isomap,lle,laplacian_eigenmaps,ltsa,manifold_charting,atlas_compatibility_transformation} unwrap the entire manifold in a single piece, so they cannot embed $n$-dimensional manifolds with non-trivial topology into $\mathbb{R}^{n}$.
Other techniques present solutions to this problem.
Extending manifold learning to work with circle-valued functions (in contrast to ordinary real line-valued functions) enables the embedding of some manifolds, like the torus, but not others, such as the sphere~\cite{circular_coordinates}.
CycleCut~\cite{cyclecut} identifies topological holes in a neighborhood graph and makes cuts until the hole is removed, so the manifold can be unwrapped with standard techniques.
A drawback is that by cutting the manifold, connectivity information is lost.

Directly learning an atlas, as done in~\cite{learning_a_manifold_as_an_atlas}, enables the reconstruction of manifolds with arbitrary topological features, without explicitly computing a global embedding.
But these charts must be linear projections, to admit fast, simultaneous optimization of domain assignments and the learned embeddings.
The Manifold Tangent Classifier~\cite{manifold_tangent_classifier} trains a contractive autoencoder for dimensionality reduction, and then extracts an atlas of linear projections to describe the local geometry of the data.
Charts with nonlinear mappings can be learned with an autoencoder~\cite{korman_atlas}, but this requires specifying the number of charts in advance, and charts may not preserve their domain's topology, as shown in Figure~\ref{fig:torus}.

\section{Methodology}%
\label{sec:methodology}

We begin with a brief overview of the theory of manifolds from the perspective of differential geometry\footnote{See~\cite{manifolds_book} for a thorough discussion of the theory of manifolds.}, before posing the problem of atlas learning and presenting our algorithm.

\subsection{Manifolds in Differential Geometry}

A $d$-dimensional smooth manifold is a set $\mathcal{M}$, such that every point $p\in\mathcal{M}$ is contained in a neighborhood $U$, with a diffeomorphism $\phi:U\to\mathbb{R}^{d}$.
The pair $(U,\phi)$ is called a coordinate chart, and a family of charts whose domains collectively cover the manifold is called an atlas.

The study of topological holes is closely related to the study of differential forms on manifolds\footnote{Differential forms are used to generalize the tools of single-variable calculus, such as integration and differentiation, to higher dimensional spaces and smooth manifolds. Explicitly, a differential form is an antisymmetric covariant tensor field.}.
The relation between forms and topological holes is quantified by the \textit{de Rham cohomology}.
For $k\in\mathbb{N}$, the $k$th de Rham cohomology group is a vector space $H^{k}_{\mathrm{dR}}(\mathcal{M})$.
A key property of these vector spaces is that for $k>0$, $\operatorname{dim}H^{k}_{\mathrm{dR}}(\mathcal{M})$ is the number of $k$-dimensional holes in the manifold, and $\operatorname{dim}H^{0}_{\mathrm{dR}}(\mathcal{M})$ is the number of connected components.
We say a space with one connected component, and no holes of any dimension, has \textit{trivial} de Rham cohomology.

Looking towards our atlas learning framework, we can see that any chart domains must have trivial de Rham cohomology.
Given two charts with trivial de Rham cohomology, if their intersection has trivial de Rham cohomology, then so does their union.
We present cases where charts cannot be combined due to this criterion in Figure~\ref{fig:why_not_combine}, and include a formal proof in the appendix.

\begin{figure}[!h]
    \centering
    \begin{subfigure}[t]{0.7\linewidth}
        \centering
        \includegraphics[width=\linewidth]{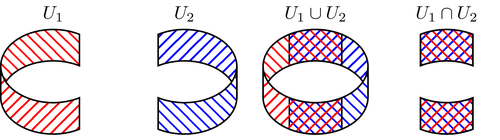}
        \caption{Two charts that together form a cylinder.}
    \end{subfigure}
    \begin{subfigure}[t]{0.7\linewidth}
        \centering
        \includegraphics[width=\linewidth]{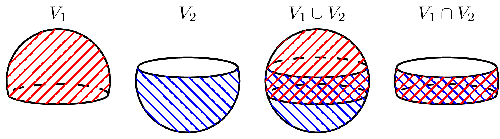}
        \caption{Two charts that together form a sphere.}
    \end{subfigure}
    \caption{This figure presents examples of cases where two charts cannot be combined without introducing a topological hole. Charts $U_{1}$ and $U_{2}$ span two halves of a cylinder, and charts $V_{1}$ and $V_{2}$ span two halves of a sphere. Our method (which only checks the intersection) is able to identify this: in the case of the cylinder, $U_{1}\cap{}U_{2}$ is disconnected, and in the case of the sphere, $V_{1}\cap{}V_{2}$ contains a hole.}
    \label{fig:why_not_combine}
\end{figure}

\subsection{Dimensionality Reduction with Manifolds and Atlases}

We pose the manifold learning problem as follows.
Given a set of $m$ points $X=\{x_{1},\ldots,x_{m}\}\in\mathbb{R}^{n}$ sampled from a $d$-dimensional smooth manifold $\mathcal{M}$, we want to construct a map $f:\mathcal{M}\to\mathbb{R}^{l}$ (with $l<n$) such that the latent information in $X$ is still present in the image under $f$.

Using manifold learning to embed $\mathcal{M}$ in $\mathbb{R}^{d}$ may fail if $\mathcal{M}$ has nontrivial de Rham cohomology; we can instead learn an atlas of charts.
We construct an atlas $\mathcal{A}=\{(U_{i},\phi_{i})\}_{i\in{}I}$, where $I$ is a finite indexing set, such that each coordinate chart preserves all latent information within its domain, and the atlas as a whole describes the global topology.

In general, it is better to construct fewer, larger charts for the simplest possible representation of the data without losing any information.
Using fewer charts makes for easier data interpretation and visualization.
Larger charts simplify the comparison of distant data points, as points in the embedding can only be directly compared within a single chart.

\subsection{Topologically-Informed Atlas Learning}

Our atlas learning approach can broadly be divided into three steps.
First, we discover local structure in the data and cover the manifold in many small charts.
We then combine these chart domains while avoiding problematic topological features.
Finally, when we can no longer combine any chart domains, we embed each chart.

\subsubsection{Neighborhood Graph Initialization}

Initially, we construct a neighborhood graph $G=(V,E)$, where each data point $x\in{}X$ corresponds to a vertex $v\in{}V$, and points are connected either to their $k$-nearest neighbors, or to all other points within a distance of $\epsilon$.
Edges are weighted by the distance between the points they connect.
This neighborhood graph describes the local geometry of the manifold, and is the starting point of many existing manifold learning algorithms~\cite{isomap,lle,laplacian_eigenmaps,ltsa}.

\subsubsection{Chart Initialization}

Throughout the chart initialization and combining process, we represent charts by their domains, as subgraphs of $G$.
To construct the original set of charts, we use iterative farthest point sampling to select a predetermined number of points from $X$. %
Each chart initially just contains that point and no edges.
Charts are then expanded to contain all of their adjacent vertices; this process is repeated until every vertex has been added to at least one chart, and then run at least one additional time to ensure adjacent charts overlap.
Thus, we have a family of domains $\mathcal{C}=\{C_{i}=(V_{i},E_{i})\}_{i\in{}I}$, such that $V=\bigcup_{i\in{}I}V_{i}$.
This step can be thought of as covering the manifold with a union of metric balls on the neighborhood graph; with sufficiently small size, we can assume these charts have trivial de Rham cohomology by the locally Euclidean property.

\subsubsection{Chart Combining}

Once we have initialized our chart domains, we begin combining chart domains.
At each step of our algorithm, we randomly select two charts $C_{i},C_{j}\in\mathcal{C}$, and compute their intersection $C_{i}\cap{}C_{j}$ as subgraphs of $G$. The charts can be combined if  $C_{i} \cap{} C_{j}$
\begin{enumerate}
    \item[1.] has precisely one connected component, and
    \item[2.] has no atomic cycles longer than a predetermined threshold $\lambda$.
\end{enumerate}
\textit{Atomic cycles} are graph cycles with no length-$n$ chords, and are an indicator of topological holes~\cite{cyclecut}.
A limitation of these criteria is they cannot detect higher-dimensional topological holes.
If the charts can be combined, we redefine $C_{i}=(V_{i}\cup{}V_{j},E_{i}\cup{}E_{j})$, and remove $C_{j}$ from $\mathcal{C}$.
Otherwise, we select a new pair of charts and try again.
This process is repeated until there are no pairs of charts that can be combined.

\subsubsection{Chart Embedding and Inverse Mapping}

For each chart, we approximate geodesic distances with shortest paths along the neighborhood graph, and then construct an embedding that best preserves these distances with metric multidimensional scaling~\cite{mds} (as in the ISOMAP algorithm).
This does not yield an explicit mapping from the embedding space to the original manifold.
Such a mapping is not a requirement of manifold learning, but is frequently used in applications.
To construct an inverse mapping, we first compute the Delaunay triangulation of the embedding~\cite{quickhull_delaunay}.
Given a point $p$ in the embedding, we find its containing simplex in the triangulation, with vertices $y_{s_{1}},\ldots,y_{s_{d+1}}$, where the $s_{k}$ subscripts denote the index of each embedding point with respect to the original data indices.
We compute the normalized barycentric coordinates $\lambda_{1},\ldots,\lambda_{d+1}$ of $p$ by solving the system of linear equations
\begin{equation}
    \left\{p=\sum_{k=1}^{d+1}\lambda_{k}y_{s_{k}};\;\;\sum_{k=1}^{d+1}\lambda_{k}=1\right\}.
\end{equation}
This solution is unique~\cite{barycentric_coords}, and $p$ is then mapped to
\begin{equation}
    \sum_{k=1}^{d+1}\lambda_{k}x_{s_{k}}.
\end{equation}
In effect, we map the simplex from the embedding to the original space.
This provides us with an arbitrary-dimension triangulation of the manifold, and allows us to map any point within the convex hull of the embedding back onto the manifold.

\section{Experiments}%
\label{sec:experiments}

We evaluate the efficacy of our topologically-informed atlas learning, and demonstrate its usefulness with compelling robotics applications.
First, we learn atlases for various challenging synthetic manifolds; we can use their well-understood topological properties to carefully evaluate our results.
Then, we use motion capture data to learn representations of a subject's gait cycle.
Finally, we learn kinematic models for articulated objects with known cyclic constraints, following the nonparametric approach described in~\cite{sturm_kinematic_models}.
We use the neighborhood trustworthiness metric~\cite{trustworthiness} to numerically evaluate our atlas learning approach, in comparison with ordinary ISOMAP and an atlas autoencoder~\cite{korman_atlas}; these results are listed in Table~\ref{tab:trustworthiness}.

\begin{table*}[h]
    \centering
    {
    \setlength{\tabcolsep}{5pt}
    \begin{tabular}{|c|c|c|c|c|c|c|c|}
        \hhline{~-------}
        \multicolumn{1}{c|}{} & \multicolumn{7}{|c|}{Trustworthiness}\\ \hline
        \multirow{2}{*}{Experiment} & \multirow{2}{*}{ISOMAP} & Atlas Autoencoder & Atlas Autoencoder & Atlas Autoencoder & Atlas Autoencoder & Our Atlas & Our Atlas\\ 
         & & (4 Charts) Worst & (4 Charts) Mean & (15 Charts) Worst & (15 Charts) Mean & Learning Worst & Learning Mean\\ \hline
         Sphere & 0.841 & 0.999 & 0.999 & 0.997 & 0.999 & 0.999 & 0.999\\ \hline
         Torus & 0.920 & 0.917 & 0.932 & 0.873 & 0.920 & 0.997 & 0.999\\ \hline
         Klein Bottle & 0.927 & 0.882 & 0.908 & 0.586 & 0.852 & 0.994 & 0.997\\ \hline
         $\operatorname{SO}(3)$ & 0.862 & 0.987 & 0.992 & 0.951 & 0.989 & 0.996 & 0.998\\ \hline
         Gait Cycle & 0.991 & 0.782 & 0.858 & 0.829 & 0.914 & 0.996 & 0.996\\ \hline
        Can Opener & 0.953 & 0.351 & 0.536 & 0.316 & 0.517 & 0.996 & 0.997\\ \hline
        Bottle Opener & 0.985 & 0.460 & 0.686 & 0.243 & 0.491 & 0.993 & 0.993\\ \hline
    \end{tabular}
    }
    \caption{For each experiment, we compare our approach, ISOMAP, and an atlas autoencoder (with differing numbers of charts). We compute the trustworthiness~\cite{trustworthiness} of each chart in the learned atlas, and list out the worst and mean values. Our approach matches or supersedes the trustworthiness of the autoencoder in every experiment.}
    \label{tab:trustworthiness}
\end{table*}

\subsection{Synthetic Manifolds}

\begin{figure}[h]
    \centering
    \begin{subfigure}[t]{0.32\linewidth}
        \centering
        \includegraphics[width=\linewidth]{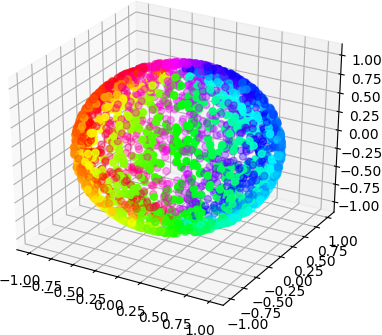}
        \caption{}
    \end{subfigure}
    \begin{subfigure}[t]{0.32\linewidth}
        \centering
        \includegraphics[width=\linewidth]{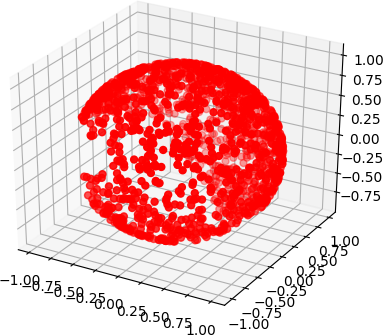}
        \caption{}
    \end{subfigure}
    \begin{subfigure}[t]{0.32\linewidth}
        \centering
        \includegraphics[width=\linewidth]{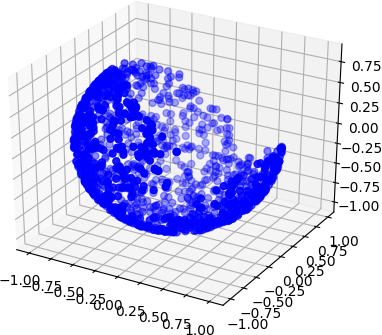}
        \caption{}
    \end{subfigure}
    \begin{subfigure}[t]{0.32\linewidth}
        \centering
        \includegraphics[width=\linewidth]{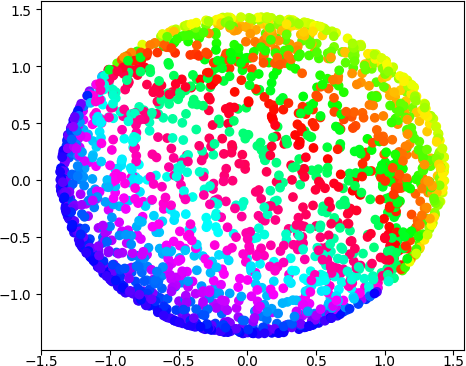}
        \caption{}
    \end{subfigure}
    \begin{subfigure}[t]{0.32\linewidth}
        \centering
        \includegraphics[width=\linewidth]{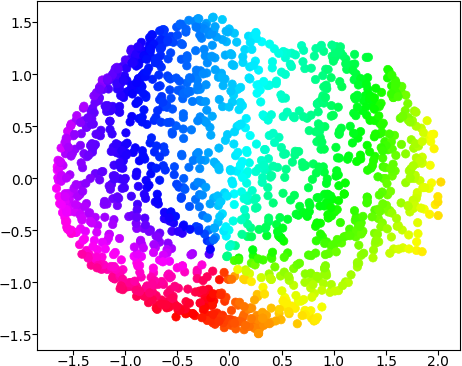}
        \caption{}
    \end{subfigure}
    \begin{subfigure}[t]{0.32\linewidth}
        \centering
        \includegraphics[width=\linewidth]{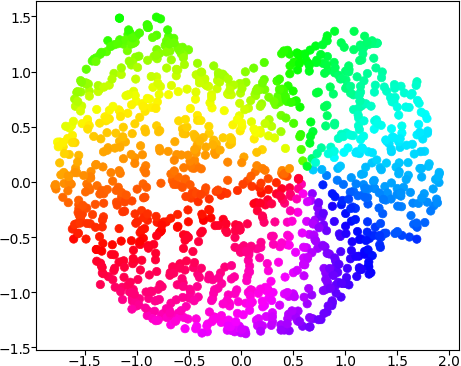}
        \caption{}
    \end{subfigure}
    \caption{Our atlas learning approach divides the sphere manifold (a) into two charts (b) and (c). The chart embeddings (e) and (f) preserve the topology of their respective domains, unlike the ISOMAP embedding (d). The sphere is parameterized by $(\cos\theta\sin\phi,\sin\theta\sin\phi,\cos\phi)$ for $\theta\in[0,2\pi),\phi\in[0,\pi]$; points are colored according to their $\theta$ parameter value.}
    \label{fig:sphere}
\end{figure}

\begin{figure*}[t]
    \centering
    \begin{subfigure}[t]{0.12\linewidth}
        \centering
        \includegraphics[width=\linewidth]{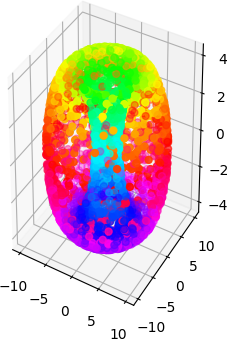}
        \caption{}
    \end{subfigure}
    \begin{subfigure}[t]{0.12\linewidth}
        \centering
        \includegraphics[width=\linewidth]{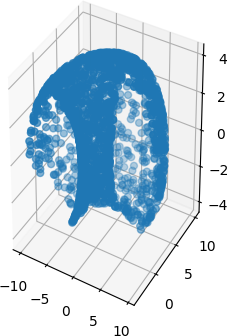}
        \caption{}
    \end{subfigure}
    \begin{subfigure}[t]{0.12\linewidth}
        \centering
        \includegraphics[width=\linewidth]{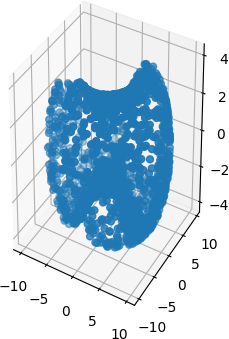}
        \caption{}
    \end{subfigure}
    \begin{subfigure}[t]{0.12\linewidth}
        \centering
        \includegraphics[width=\linewidth]{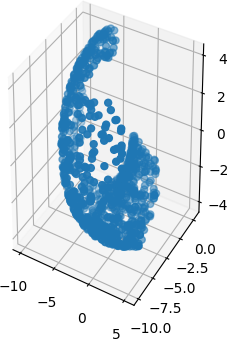}
        \caption{}
    \end{subfigure}
    \begin{subfigure}[t]{0.12\linewidth}
        \centering
        \includegraphics[width=\linewidth]{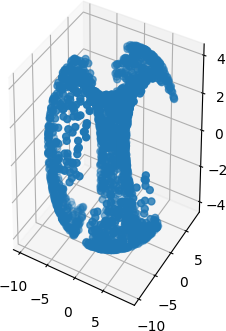}
        \caption{}
    \end{subfigure}
    \begin{subfigure}[t]{0.12\linewidth}
        \centering
        \includegraphics[width=\linewidth]{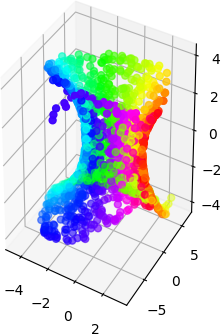}
        \caption{}
    \end{subfigure}
    \break
    \begin{subfigure}[t]{0.12\linewidth}
        \centering
        \includegraphics[width=\linewidth]{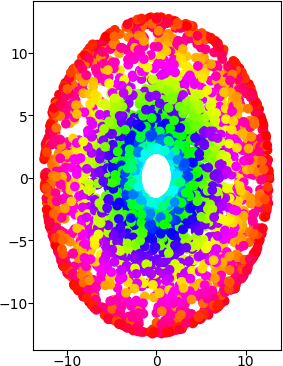}
        \caption{}
    \end{subfigure}
    \begin{subfigure}[t]{0.12\linewidth}
        \centering
        \includegraphics[width=\linewidth]{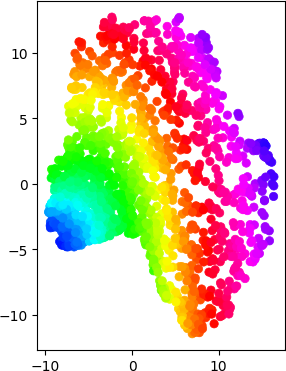}
        \caption{}
    \end{subfigure}
    \begin{subfigure}[t]{0.12\linewidth}
        \centering
        \includegraphics[width=\linewidth]{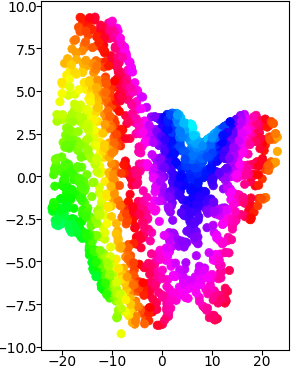}
        \caption{}
    \end{subfigure}
    \begin{subfigure}[t]{0.12\linewidth}
        \centering
        \includegraphics[width=\linewidth]{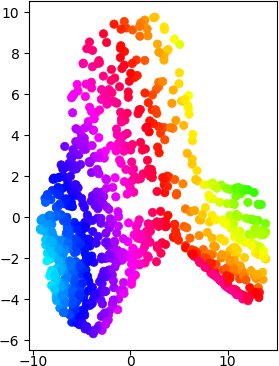}
        \caption{}
    \end{subfigure}
    \begin{subfigure}[t]{0.12\linewidth}
        \centering
        \includegraphics[width=\linewidth]{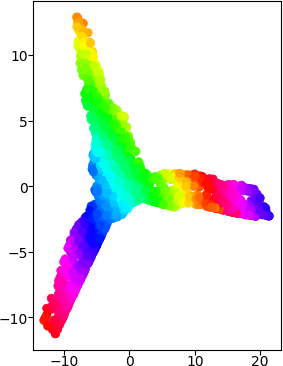}
        \caption{}
    \end{subfigure}
    \begin{subfigure}[t]{0.12\linewidth}
        \centering
        \includegraphics[width=\linewidth]{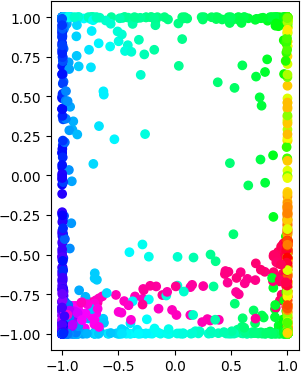}
        \caption{}
    \end{subfigure}
    \caption{Our atlas learning approach divides the torus manifold (a) into four charts (b) - (e). The ISOMAP embedding of the whole dataset (g) does not preserve local neighborhoods, but the individual chart embeddings (h) - (k) do. The torus is parameterized by $((6+4\cos\theta)\cos\phi,(6+4\sin\theta)\sin\phi,4\sin\theta)$ for $\theta,\phi\in[0,2\pi)$; points are colored according to their $\phi$ parameter. We also present one of the charts from an atlas autoencoder. Its domain (f) has a topological hole, which leads to an inaccurate embedding (l). (Points in (f) and (l) are colored by the $\theta$ parameter.)} %
    \label{fig:torus}
    
\end{figure*}

Our technique covers a sphere with two roughly-hemispherical charts, producing accurate embeddings, as shown in Figure~\ref{fig:sphere}.
ISOMAP squashes the sphere into a disk, and hence fails to preserve local neighborhoods.
Our approach covers the torus with four charts, producing accurate embeddings, as shown in Figure~\ref{fig:torus}.
ISOMAP squashes the torus into an annulus, and hence fails to preserve local neighborhoods.
The atlas autoencoder (set to create four charts) constructs a chart domain with a hole in it, producing an inaccurate embedding.
We also examine data sampled from the Klein bottle and the lie group $\operatorname{SO}(3)$; our approach's superior performance is demonstrated in Table~\ref{tab:trustworthiness}.


\subsection{Representing Human Motion}

Human motion signals are important because they can help robots perceive humans in the world around them, and they can be used to help robots inform their own motion patterns~\cite{jenkins2007tracking,dasgupta1999making}. This data is often composed of high-dimensional signals corresponding to the degrees of freedom of the human body. Extensive studies have employed manifold learning to reduce the dimensionality of these data and represent motion for tasks like activity modeling and motion reconstruction \cite{learning_a_manifold_as_an_atlas, schwarz2010amdo}.

\cite{learning_a_manifold_as_an_atlas} previously employed atlas learning to perform motion reconstruction. However, their method constructs an atlas of linear projections, requiring a large number of charts to describe the manifold. \cite{schwarz2010amdo} constructed activity models for multiple types of activities, such as clapping, walking, golfing, and waving.
Our goal was to construct a low-dimensional representation of a subject's gait cycle.
Because of the cyclic nature of this motion, atlas learning is required to learn an accurate one-dimensional embedding.

\begin{figure}[h]
    \centering
    \parbox{\linewidth}{
        \centering
        \hfill
        \begin{subfigure}[t]{\linewidth}
            \centering
            \includegraphics[width=\linewidth]{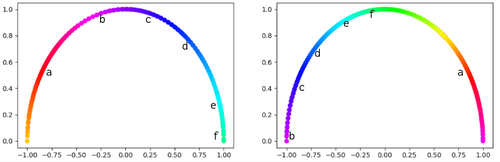}
            \caption*{}
        \end{subfigure}
        \hfill
    }
    \hfill
    \parbox{\linewidth}{
        \centering
        \hfill
        \begin{subfigure}[t]{0.15\linewidth}
            \centering
            \includegraphics[width=0.7\linewidth,height=1.5cm]{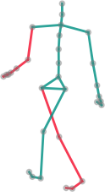}
            \caption{}
        \end{subfigure}
        \hfill
        \begin{subfigure}[t]{0.15\linewidth}
            \centering
            \includegraphics[width=0.7\linewidth,height=1.5cm]{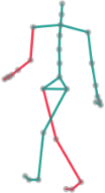}
            \caption{}
        \end{subfigure}
        \hfill
        \begin{subfigure}[t]{0.15\linewidth}
            \centering
            \includegraphics[width=0.7\linewidth,height=1.5cm]{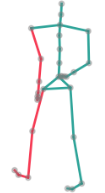}
            \caption{}
        \end{subfigure}
        \hfill
        \begin{subfigure}[t]{0.15\linewidth}
            \centering
            \includegraphics[width=0.7\linewidth,height=1.5cm]{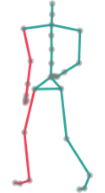}
            \caption{}
        \end{subfigure}
        \hfill
        \begin{subfigure}[t]{0.15\linewidth}
            \centering
            \includegraphics[width=0.7\linewidth,height=1.5cm]{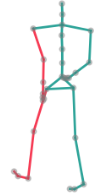}
            \caption{}
        \end{subfigure}
        \hfill
        \begin{subfigure}[t]{0.15\linewidth}
            \centering
            \includegraphics[width=0.7\linewidth,height=1.5cm]{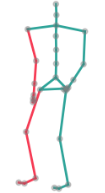}
            \caption{}
        \end{subfigure}
    }
    \caption{Learned embedding for a walking sequence from the CMU Mocap dataset. On the left are a pair of one-dimensional coordinate charts comprising this embedding, parameterized on a semicircle, with the color corresponding to the timestep in the sequence. (a) - (f) are poses in the walking sequence that lie in the overlapping region of both charts. Their corresponding parameterized embedding locations are annotated on the charts.}
    \label{fig:walking_manifold_seq}
\end{figure}

We embed data from the CMU Mocap Dataset\footnote{\texttt{\href{http://mocap.cs.cmu.edu/}{http://mocap.cs.cmu.edu/}}}, which contains 31 joint poses per frame. Given an action sequence, we normalize the joints in each individual frame and embed it using our atlas learning method.
Figure \ref{fig:walking_manifold_seq} depicts the learned embedding of a subject walking; the two charts are an accurate representation of their gait cycle.

\subsection{Learning Kinematic Models} 

Articulated objects are composed of multiple rigid links, connected together by nonrigid joints.
Such objects inherently have more degrees of freedom (beyond the 6DoF pose associated to rigid objects), presenting an added challenge for robot perception and manipulation tasks.
Many algorithms for such tasks directly or indirectly utilize kinematic models~\cite{paolillo2018interlinked,pavlasek2020parts,capitanelli2018manipulation}, so the unsupervised learning of kinematic models is a subject of great interest in robotics.

Some approaches to modeling articulated objects explicitly describe the joints as either rotational or prismatic~\cite{reinforcement_learning_articulated_objects,kinematic_chain_building,huang2012occlusion}.
To handle irregular joints or hidden kinematic constraints, Sturm \etal{} use manifold learning to build non-parametric models~\cite{sturm_kinematic_models}.
Given a sequence of object states, they construct a $d$-dimensional embedding, where $d$ is the number of degrees of freedom.
However, rotational constraints with no joint limits appear as holes in the manifold, necessitating atlas learning.

We follow the non-parametric approach of Sturm \etal{} \cite{sturm_kinematic_models} to construct our kinematic models.
We place AprilTags~\cite{apriltags} (visual fiducial markers) on the different parts of the articulated object.
Given a video of the object being manipulated, in each image, we detect the pose of each tag and transform them into a local object coordinate frame.
We concatenate the positions of each tag corner to use as the input data for atlas learning.
For each chart, the inverse mapping allows us to construct the tag poses (and thus the link poses) for any point in the convex hull of the embedding.

\begin{figure*}[h]
    \centering
    \parbox{0.3\linewidth}{
        \centering
        \hfill
        \begin{subfigure}[t]{\linewidth}
            \centering
            \includegraphics[width=\linewidth]{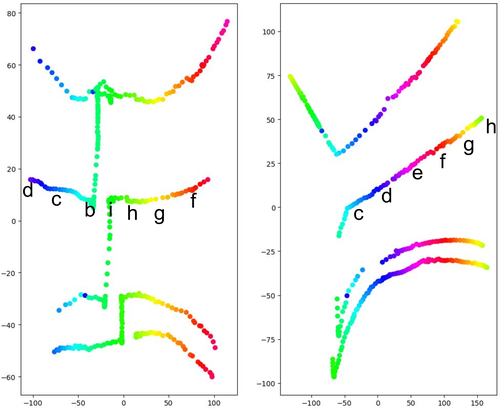}
            \caption{}
        \end{subfigure}
        \hfill
    }
    \parbox{0.595\linewidth}{
        \centering
        \hfill
        \begin{subfigure}[t]{0.24\linewidth}
            \centering
            \includegraphics[width=\linewidth]{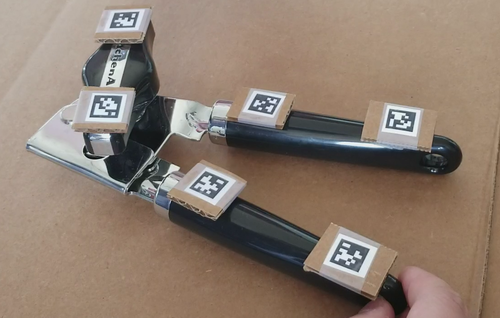}
            \caption{}
        \end{subfigure}
        \hfill
        \begin{subfigure}[t]{0.24\linewidth}
            \centering
            \includegraphics[width=\linewidth]{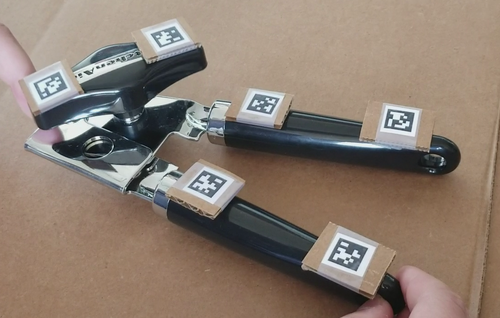}
            \caption{}
        \end{subfigure}
        \hfill
        \begin{subfigure}[t]{0.24\linewidth}
            \centering
            \includegraphics[width=\linewidth]{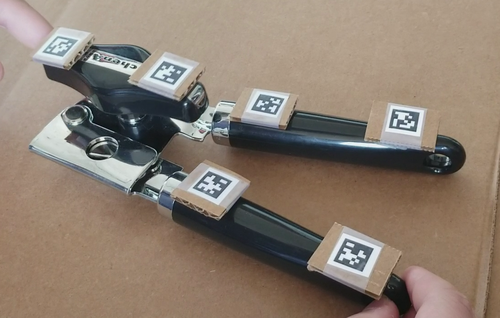}
            \caption{}
        \end{subfigure}
        \hfill
        \begin{subfigure}[t]{0.24\linewidth}
            \centering
            \includegraphics[width=\linewidth]{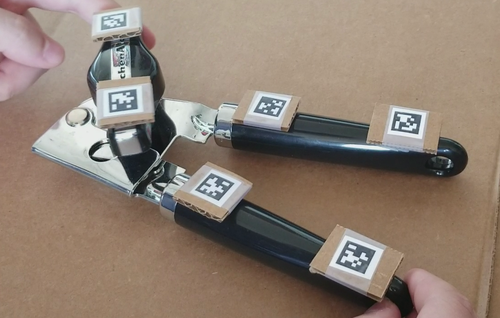}
            \caption{}
        \end{subfigure}
        \hfill
        \begin{subfigure}[t]{0.24\linewidth}
            \centering
            \includegraphics[width=\linewidth]{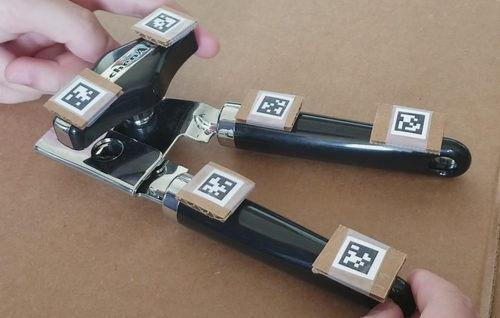}
            \caption{}
        \end{subfigure}
        \hfill
        \begin{subfigure}[t]{0.24\linewidth}
            \centering
            \includegraphics[width=\linewidth]{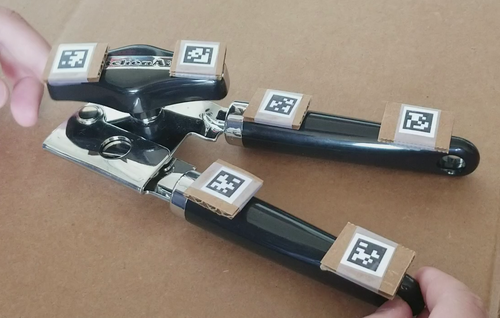}
            \caption{}
        \end{subfigure}
        \hfill
        \begin{subfigure}[t]{0.24\linewidth}
            \centering
            \includegraphics[width=\linewidth]{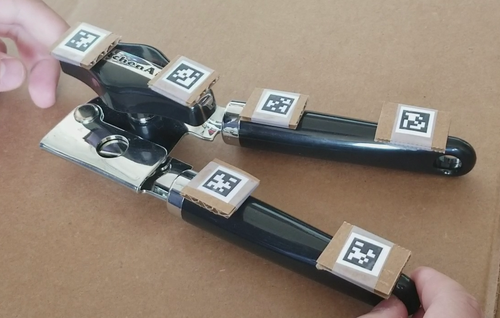}
            \caption{}
        \end{subfigure}
        \hfill
        \begin{subfigure}[t]{0.24\linewidth}
            \centering
            \includegraphics[width=\linewidth]{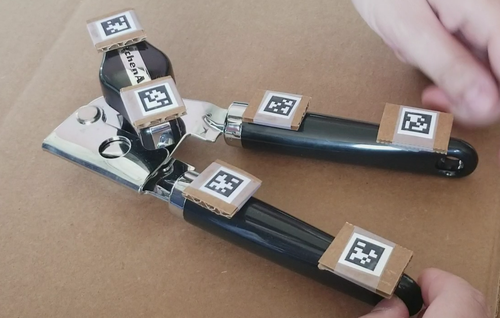}
            \caption{}
        \end{subfigure}
    }
    \caption{The learned nonparametric kinematic model for the can opener. (a) contains embeddings of the two coordinate charts, colored according to the angle made by the smaller handle. Moving vertically in the embedding corresponds to opening the large handles, and moving horizontally across the two charts encodes the rotation of the smaller handle. The charts are annotated with the embedding location of input images, (b) - (i), where the handle was twisted in a full circle.}
    \label{fig:can_opener_seq_1}
\end{figure*}

The first object we build a kinematic model of is a can opener.
It has two degrees of freedom: the large handles used to grip the can, and the smaller, twist handle to open the can.
Because the twist handle can freely rotate across 360 degrees, but the larger handle has a limit to its motion, the manifold has a cylindrical structure in the high-dimensional space: the twist handle corresponds to moving around the circle, and the large handle corresponds to moving along the cylinder.
This cylindrical structure necessitates atlas learning to construct a low-dimensional representation of the data.
We include the learned embedding, along with the embedded location of several input images, in Figure~\ref{fig:can_opener_seq_1}.
Moving vertically in this embedding corresponds to opening or closing the large handles, and moving horizontally across the two charts encodes the rotation of the smaller handle.

\begin{figure*}[h]
    \centering
    \parbox{0.17\linewidth}{
        \centering
        \begin{subfigure}[t]{\linewidth}
            \centering
            \includegraphics[width=\linewidth]{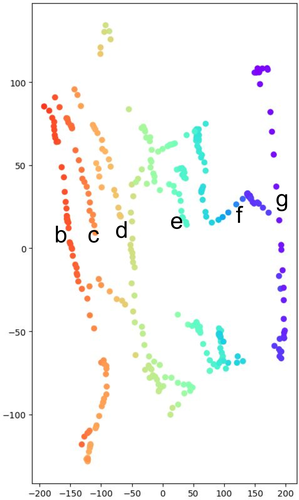}
            \caption{}
        \end{subfigure}
    }
    \parbox{0.6\linewidth}{
        \centering
        \hfill
        \begin{subfigure}[t]{0.32\linewidth}
            \centering
            \includegraphics[width=\linewidth]{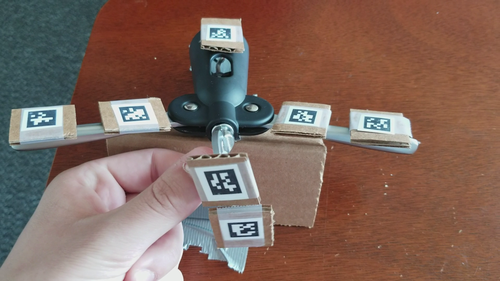}
            \caption{}
        \end{subfigure}
        \hfill
        \begin{subfigure}[t]{0.32\linewidth}
            \centering
            \includegraphics[width=\linewidth]{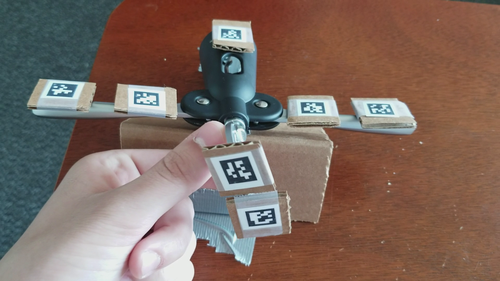}
            \caption{}
        \end{subfigure}
        \hfill
        \begin{subfigure}[t]{0.32\linewidth}
            \centering
            \includegraphics[width=\linewidth]{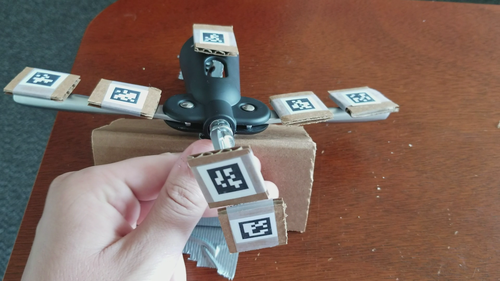}
            \caption{}
        \end{subfigure}
        \hfill
        \begin{subfigure}[t]{0.32\linewidth}
            \centering
            \includegraphics[width=\linewidth]{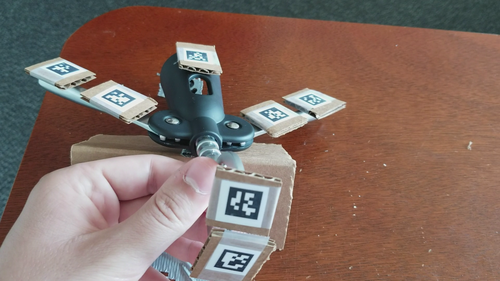}
            \caption{}
        \end{subfigure}
        \hfill
        \begin{subfigure}[t]{0.32\linewidth}
            \centering
            \includegraphics[width=\linewidth]{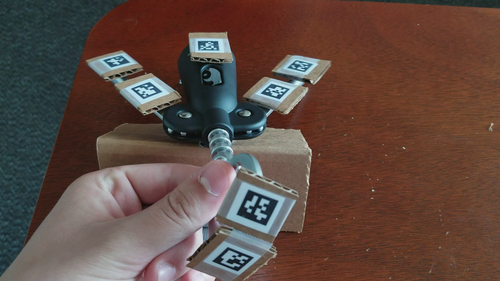}
            \caption{}
        \end{subfigure}
        \hfill
        \begin{subfigure}[t]{0.32\linewidth}
            \centering
            \includegraphics[width=\linewidth]{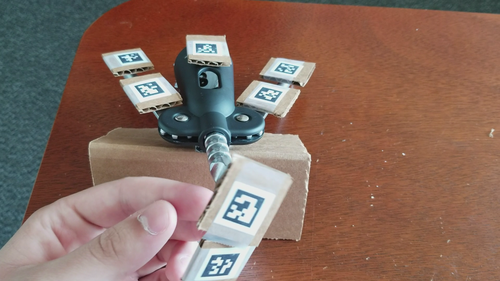}
            \caption{}
        \end{subfigure}
    }
    \caption{Part of the learned nonparametric kinematic model for the bottle opener. (c) contains the embedding of one of the coordinate charts, colored according to the angle of the side lever arms. Moving horizontally in the embedding corresponds to sliding the corkscrew, as well as the linked rotation of the side handles. The chart is annotated with the embedding location of several input images, (d) - (i), encompassing the range of motion of the corkscrew and side handles.}
    \label{fig:corkscrew_seq_1}
\end{figure*}

The next object we build a kinematic model of is a corkscrew bottle opener.
This object has four individual parts, but only two degrees of freedom.
The corkscrew itself can rotate freely relative to the base, and also translate along the rotation axis.
The two lever arms can rotate, but their motion is controlled by the prismatic motion of the corkscrew.
This motion is described by the kinematic model, as shown in Figure~\ref{fig:corkscrew_seq_1}.
The chart is colored according to the position of the angle of the lever arms, and the embedding location of several input images is shown.

\section{Conclusion}%
\label{sec:conclusion}

In this paper, we have presented a new technique for learning an atlas of coordinate charts, in order to perform dimensionality reduction of manifold data.
Our approach is able to partition the data into domains which each contain no holes, ensuring that each chart can be accurately embedded.
By learning atlases for challenging computer-generated manifolds, reconstructing human motion capture data, and learning kinematic models for articulated objects, we have demonstrated the efficacy of our approach.

For certain tasks, using a different technique to look for topological holes may be beneficial.
Although we chose to search for atomic cycles, there are other ways to identify holes in neighborhood graphs~\cite{holes_and_antiholes,finding_large_holes}.
Also, higher-dimensional holes may not be detected as cycles in neighborhood graphs, and the de Rham cohomology does not capture all topological properties of a manifold (for example, $\mathbb{R}P^{2}$ has trivial de Rham cohomology, but is not homeomorphic to a Euclidean space).
Detecting such topological features is an open question, but tools of topological data analysis, such as persistent homology~\cite{persistent_homology}, are a promising starting point.

\section*{Appendix}
\label{sec:appendix}
\begin{figure}[!h]
	\centering
	\includegraphics[width=0.65\linewidth]{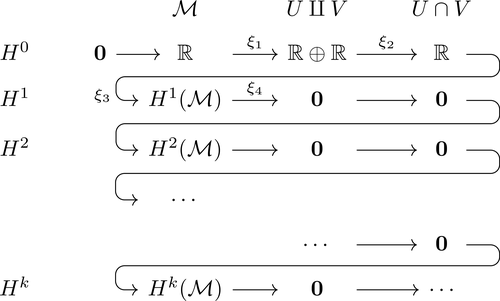}
	\caption{The Mayer-Vietoris sequence for our proof, with initially-known values filled in.}
	\label{fig:mayer-vietoris}
\end{figure}

In this section, we prove that we only need to examine the intersection of two charts to check if they can be combined.
We first present some necessary additional background.
Given a sequence of vector spaces $\{V_{i}\}_{i\in\mathbb{Z}}$, a sequence of linear maps $f_{i}:V_{i}\to{}V_{i+1}$ is called exact if, $\forall{}i$, $\operatorname{im}f_{i}=\operatorname{ker}f_{i+1}$.
We use the Mayer-Vietoris sequence, which is an exact sequence on the de Rham cohomology groups of a manifold that is made of two overlapping components.
This sequence is used to compute the topological properties of a manifold, in a process called ``diagram chasing''.
Notationally, $\mathbf{0}$ denotes the zero vector space, and $U\amalg{}V$ is the disjoint union of $U$ and $V$.
Let $\mathcal{M}=U\cup{}V$ be a smooth manifold. If $U$, $V$, and $U\cap{}V$ have trivial de Rham cohomology, then so does $\mathcal{M}$. \emph{Proof}: Because $U$, $V$, and $U\cap{}V$ have trivial de Rham cohomology, they must each be connected, so $\mathcal{M}$ is connected~\cite[p.~333]{manifolds_book}, and $H^{0}_{\mathrm{dR}}(\mathcal{M})\cong\mathbb{R}$. Well,
\begin{equation}
    H^{k}_{\mathrm{dR}}(U)\cong{}H^{k}_{\mathrm{dR}}(V)\cong{}H^{k}_{\mathrm{dR}}(U\cap{}V)\cong\left\{\begin{array}{ll}\mathbb{R} & k=0\\ \mathbf{0} & k>0\end{array}\right.
\end{equation}
This means the cohomology groups of $U\amalg{}V$ are
\begin{equation}
    H^{k}_{\mathrm{dR}}(U\amalg{}V)\cong\left\{\begin{array}{ll}\mathbb{R}\oplus\mathbb{R} & k=0\\ \mathbf{0} & k>0\end{array}\right.
\end{equation}
We organize the Mayer-Vietoris sequence in tabular form, and fill in known values, as shown in Figure~\ref{fig:mayer-vietoris}.

We then compute the remaining de Rham cohomology groups for $\mathcal{M}$.
If the sequence $\mathbf{0}\overset{f_{k}}{\to}H^{k}_{\mathrm{dR}}(\mathcal{M})\overset{g_{k}}{\to}\mathbf{0}$ is exact, then $\operatorname{dim}(\operatorname{im}(f_{k}))=0$, so $\operatorname{dim}(\operatorname{ker}(g_{k}))=0$. Because $g_{k}$ maps everything to $0$, $\operatorname{ker}(g_{k})=H^{k}_{\mathrm{dR}}(\mathcal{M})=\mathbf{0}$, $\forall k>1$.

What remains is computing $H^{1}_{\mathrm{dR}}(\mathcal{M})$.
By exactness, $\operatorname{ker}(\xi_{1})=\mathbf{0}$, so $\xi_{1}$ is injective, so $\operatorname{dim}(\operatorname{im}(\xi_{1}))=1$.
$\operatorname{im}(\xi_{1})=\operatorname{ker}(\xi_{2})$, so $\operatorname{dim}(\operatorname{ker}(\xi_{2}))=1$.
By rank-nullity, $\operatorname{dim}(\operatorname{im}(\xi_{2}))=1$, so $\xi_{2}$ is surjective.
This means $\operatorname{im}(\xi_{2})=\mathbb{R}$, so $\operatorname{ker}(\xi_{3})=\operatorname{im}(\xi_{2})=\mathbb{R}$, so $\operatorname{dim}(\operatorname{im}(\xi_{3}))=0$, by rank-nullity.
But $\operatorname{ker}(\xi_{4})=\operatorname{im}(\xi_{3})$, so $\operatorname{dim}(\operatorname{ker}(\xi_{4}))=0$.
And $\operatorname{dim}(\operatorname{im}(\xi_{4}))=0$, so by rank nullity, $\operatorname{dim}(H^{1}_{\mathrm{dR}}(\mathcal{M}))=0$.
Thus $\mathcal{M}$ has trivial de Rham cohomology.

\section*{Acknowledgment}
\label{sec:acknowledgment}
The authors thank Professor Alejandro Uribe (University of Michigan) for his invaluable guidance and direction in developing the initial theory behind this work.

\bibliographystyle{IEEEtran}
\bibliography{root.bib}

\end{document}